%% file: main.tex
\documentclass[conference]{IEEEtran}
\IEEEoverridecommandlockouts
\usepackage{cite}
\usepackage{amsmath,amssymb,amsfonts}
\usepackage{algorithm}
\usepackage{algpseudocode}

\algrenewcommand\algorithmicfor{\textbf{for}}
\algrenewcommand\algorithmicdo{\textbf{do}}
\algrenewcommand\algorithmicif{\textbf{if}}
\algrenewcommand\algorithmicthen{\textbf{then}}
\algrenewcommand\algorithmicend{\textbf{end}}

\algrenewcommand\algorithmicrequire{\textbf{input:}}
\algrenewcommand\algorithmicensure{\textbf{output:}}

\newcommand\yl[1]{{\color{blue}{#1}}}

\usepackage{graphicx}
\usepackage{textcomp}
\usepackage{xcolor}
\def\BibTeX{{\rm B\kern-.05em{\sc i\kern-.025em b}\kern-.08em
    T\kern-.1667em\lower.7ex\hbox{E}\kern-.125emX}}
\begin{document}


\title{\LARGE \textbf{
Feasibility Restoration under Conflicting STL Specifications\\ with Pareto-Optimal Refinement 
}\\
{\large }


\author{\large Tianhao Wu$^1$, Yiwei Lyu$^2$}
\thanks{ \yl{}

 $^1$ The author is with the Department of Computer Science, University of Southern California. Email: wutianha@usc.edu
 
 $^2$ The author is with the Department of Computer Science and Engineering, Texas A\&M University. Email: yiweilyu@tamu.edu}
}

\maketitle


\begin{abstract}
Signal Temporal Logic (STL) is expressive formal language that specifies spatio-temporal requirements in robotics. Its quantitative robustness semantics can be easily integrated with optimization-based control frameworks. However, STL specifications may become conflicting in real-world applications, where safety rules, traffic regulations, and task objectives can be cannot be satisfied together. In these situations, traditional STL-constrained Model Predictive Control (MPC) becomes infeasible and default to conservative behaviors such as freezing, which can largely increase risks in safety-critical scenarios.
In this paper, we proposes a unified two-stage framework that first restores feasibility via minimal relaxation, then refine the feasible solution by formulating it as a value-aware multi-objective optimization problem. Using $\varepsilon$-constraint method, we approximate the Pareto front of the multi-objective optimization, which allows analysis of tradeoffs among competing objectives and counterfactual analysis of alternative actions.
We demonstrate that the proposed approach avoids deadlock under conflicting STL specifications and enables interpretable decision-making in safety-critical applications by conducting a case study in autonomous driving.

\end{abstract}


\input{introduction}

\input{preliminaries}

\input{method2}

\input{experiments}

\input{conclusion}

\bibliographystyle{IEEEtran}
\bibliography{refs}


\end{document}

%% file: introduction.tex
\section{Introduction}

Temporal logic (TL) is a class of formal specification languages that describes how system should behave over time \cite{rescher2012temporal}. In safety-critical robotics, TL encodes high-level task goals and safety rules so they can be systematically verified and enforced. For example, an autonomous vehicle specification such as, “if a stop sign is detected, the vehicle must come to a complete stop”, can be formally written as $\varphi_{\text{stop}} \mathbf{G}\Big( \text{stop\_detected} \rightarrow\; \mathbf{F}_{[0,T]}\big(v_{\text{ego}}=0 \big)\Big)
$. With this ability to unambiguously characterize spatio-temporal properties, Linear Temporal Logic (LTL) \cite{pnueli1977temporal}, Metric Temporal Logic (MTL) \cite{koymans1990specifying}, and Computation Tree Logic (CTL) \cite{clarke1981design} are widely used to specify tasks and safety requirements in safety-critical robotics. Prior work has applied temporal logics to encode task specifications in risk-aware autonomous driving \cite{qi2025, arfvidsson2024towards}, safe human-robot interaction \cite{yu2024}, and UAV trajectory optimization \cite{chen2025}.

A central appeal of temporal logic is that it provides a principled interface between symbolic requirements and continuous robot dynamics. In automaton-based pipelines, a temporal-logic formula is compiled into a finite-state monitor (or automaton) that tracks progress toward satisfaction \cite{baier2008principles}. The controller then plans in a product space of robot states and automaton states to ensure the monitor remains in accepting behaviors \cite{luo2022, bergeron2024comp, chen2024fast}. In optimization-based pipelines, temporal requirements are translated into constraints and objectives inside trajectory optimization or Model Predictive Control (MPC), allowing the controller to satisfy logical rules while optimizing performance criteria such as efficiency, smoothness, and task completion time \cite{raman2014, gautam2025rrt, ren2024ltl}. This latter route is especially compelling when the logic admits quantitative semantics that can be optimized directly. 

Recently, Signal Temporal Logic (STL) has drawn growing attention because it can reason over real-valued, continuous-time signals and provide a robustness score that measures the extent to which the specification is satisfied. This quantitative semantics makes STL particularly easily integrable with optimization-based control, where STL formulas can be converted into constraints or incorporated as robustness-maximizing objectives \cite{verhagen2023, wang2024}. 

However, strictly enforcing all STL constraints can make the online optimization infeasible when specifications directly conflict, which often triggers a conservative fallback that “freezes” the robot. In an autonomous driving setting, for example, a conflict can arise from simultaneously requiring yield to an approaching ambulance, never cross the double-yellow centerline, and maintain a minimum clearance to nearby obstacles and vehicles. In a narrow or partially blocked lane, there may be no maneuver that satisfies all three requirements at once, so the controller finds no feasible solution and defaults to stopping or waiting, effectively becoming a road obstruction in time-critical situations. This type of deadlock has appeared in real-world AV deployments, including reports about robotaxis blocking an ambulance in San Francisco \cite{anguiano2023} and a similar event in Austin recently \cite{cobler_waymo_2026}, which illustrates why feasibility restoration under competing specifications is critical.

A substantial body of work has studied STL-constrained control and robustness-based synthesis. Early approaches embed STL constraints into mixed-integer programs within MPC, guaranteeing satisfaction whenever the induced problem is solvable, but do not address cases when specifications conflict and no jointly satisfying solution exists. Robustness-maximization methods improve satisfaction margins through quantitative STL semantics, yet they typically still assume that the overall specification set can be met \cite{gilpin2020}. More recent work introduces temporal relaxation metrics \cite{buyukkocak2022}, but time-domain adjustment alone does not capture partial violations that arise in the state space. \cite{cardona2023} pursue partial satisfaction via lexicographic optimization, but the required fixed priority ordering prevents context-dependent tradeoffs among requirements. In summary, we still lack a unified optimization-based control framework that can (i) systematically handle conflicting STL specifications online, and (ii) make the inevitable tradeoffs among competing requirements explicit and adjustable. This gap matters in safety-critical autonomy, where conflicts are unavoidable and the controller must make principled compromises rather than freezing.

In this work, we develop a two-stage framework for resolving conflicts among STL specifications in optimization-based control: The first stage restores MPC feasibility by minimally relaxing negotiable requirements while strictly enforcing non-negotiable safety constraints; the second stage then refines the recovered solution by explicitly searching over tradeoffs among competing objectives to surface a set of value-consistent alternatives. Our \textbf{main contributions} are: 
(1) We present an infeasibility-recovery mechanism based on minimal $L_1$-norm relaxation that restores MPC feasibility with the smallest necessary specification violation, producing least-deviation behavior rather than conservative ``freezing''  fallback. 
(2) With the recovered feasible baseline, we propose a value-aware refinement stage that approximates the Pareto front via the $\varepsilon$-constraint method, enabling systematic exploration of performance tradeoffs and counterfactual reasoning about how alternative actions would change each objective. (3) We demonstrate the effectiveness of the proposed method in a case study of autonomous driving with conflicting specifications.

%% file: preliminaries.tex
\section{Preliminaries on Signal Temporal Logic (STL)}


STL was first introduced in \cite{maler2004}\ to specify temporal properties of real-valued signals, where signal $S$ is a function $S : \tau \rightarrow \mathbb{R}^n$, and $\tau \subseteq \mathbb{R}_{\ge 0}$ is the time domain. It is a formal language with a defined syntax and semantics, which specify what kinds of STL formulae are allowed and how to interpret them respectively. In the next two sections, we formally introduce the standard syntax and semantics of STL.
\subsubsection{Syntax}

Let  $I = [a,b] \subseteq \tau$ denote the time interval $\{t \in \tau \mid a \le t \le b\}$, then a STL formula $\varphi$ is defined recursively as
\begin{equation}
\varphi ::= \top \mid \mu \mid \neg \varphi \mid 
\varphi_1 \wedge \varphi_2 \mid 
\mathbf{F}_I \varphi \mid 
\mathbf{G}_I \varphi
\end{equation}

The syntax specifies that an STL formula $\varphi$ can be a boolean constant $\top$ (true), 
a predicate $\mu$ of the form
$\mu := l(S(t)) \ge 0$
with $l : \mathbb{R}^n \to \mathbb{R}$ Lipschitz continuous, or a composition of STL formulae $\varphi_1, \varphi_2$ using logical operators $\neg$, $\wedge$
(e.g. $\neg \varphi_1 \wedge \varphi_2$) and temporal operators $\mathbf{F}_I$, $\mathbf{G}_I$
(e.g. $\mathbf{G}_I (\varphi_2 \wedge \varphi_2)$). 
The temporal operator $\mathbf{F}_I$ (eventually) specifies that there exists some $t' \in t+I$ such that $\varphi$ is true at $t'$. $\mathbf{G}_I$ (always) specifies that $\varphi$ holds true for all $t' \in t+I$.
For example,
$\mathbf{F}_{[0,10]} \mathbf{G}_{[0,5]} \big( S \ge \alpha \big)$
is interpreted as: in the next 10 time steps, there exists some time $t \in [0,10]$ such that $S \ge \alpha$ always holds true between $[t, t+5]$.

\subsubsection{Semantics}
STL has two semantics: qualitative (boolean) and quantitative (robustness). Qualitative semantics determines whether a signal $S$ satisfies a specification $\varphi$ at time $t$. We denote $(S,t) \models \varphi$ as satisfaction and $(S,t) \not\models \varphi$ as violation.
Quantitative semantics measures the extent to which $S$ satisfies $\varphi$.
The robustness score, denoted as $\rho(\varphi,S,t) \in \mathbb{R}$, is defined recursively as
\begin{equation}
\begin{aligned}
\rho(\mu,S,t) &= l(S(t)), \\
\rho(\neg \varphi,S,t) &= -\rho(\varphi,S,t), \\
\rho(\varphi_1 \wedge \varphi_2,S,t) 
&= \min\!\big(\rho(\varphi_1,S,t), \rho(\varphi_2,S,t)\big), \\
\rho(\varphi_1 \vee \varphi_2,S,t) 
&= \max\!\big(\rho(\varphi_1,S,t), \rho(\varphi_2,S,t)\big), \\
\rho(\mathbf{G}_I \varphi,S,t) 
&= \inf_{t' \in t+I} \rho(\varphi,S,t'), \\
\rho(\mathbf{F}_I \varphi,S,t) 
&= \sup_{t' \in t+I} \rho(\varphi,S,t').
\end{aligned}
\end{equation}

An important property of robustness is that it is sound and complete with respect to the Boolean semantics, namely,
\begin{equation}
\rho(\varphi,S,t) \ge 0 \iff (S,t) \models \varphi.
\label{eq:stl_robustness}
\end{equation}
From the equivalence, positive robustness indicates satisfaction of $\varphi$, and negative robustness indicates violation of $\varphi$.

%% file: method2.tex
\section{Methodology}

\subsection{Problem Formulation}
We consider an MPC problem over a horizon $T$ with system dynamics
$x_{t+1} = f(x_t,u_t)$ for $t=0,\dots,T-1$,
and admissible inputs $u_t \in \mathcal{U}$. Let $\Phi=\Phi_H\cup\Phi_S$ denote a set of STL specifications, partitioned into non-negotiable formulas $\Phi_H$ (e.g., robot physical limits) and negotiable formulas $\Phi_S$ (e.g., rules and objectives that may need to be compromised under conflict). For each STL formula $\varphi\in\Phi$, let $\rho_\varphi(x_{0:T})$ denote its robustness score, where $\rho_\varphi(x_{0:T})\ge 0$ indicates satisfaction.

A standard STL-constrained MPC optimizes a nominal objective while enforcing $\rho_\varphi(x_{0:T})\ge 0$ for all $\varphi\in\Phi$. In realistic deployments, however, the specification set $\Phi$ can become internally inconsistent at runtime, so that no control sequence can satisfy all formulas simultaneously. This motivates two key questions:
\begin{enumerate}
    \item \textbf{Feasibility restoration:} When $\Phi$ is conflicting under the current context, how can we restore feasibility by minimally relaxing only negotiable specifications while keeping $\Phi_H$ strict?
    \item \textbf{Tradeoff characterization:} Once feasibility is restored, how can we evaluate and represent the induced tradeoffs among competing consequence objectives across alternative feasible conflict resolutions?
\end{enumerate}

\subsection{Minimal-Relaxation Feasibility Restoration}
\label{subsec:stage1}

We begin with the failure mode where STL-constrained MPC becomes infeasible under the current context. Consider the nominal STL-constrained MPC
\begin{equation}
\begin{aligned}
\min_{u_{0:T-1}} \quad & J(x_{0:T},u_{0:T-1}) \\
\text{s.t.}\quad
& x_{t+1} = f(x_t,u_t), \quad t=0,\dots,T-1,\\
& u_t \in \mathcal{U}, \quad t=0,\dots,T-1,\\
& \rho_{\varphi}(x_{0:T}) \ge 0, \quad \forall \varphi\in\Phi,
\end{aligned}
\label{eq:stl_mpc}
\end{equation}
which admits a feasible solution only when the STL constraints are jointly satisfiable. Stage~1 restores feasibility by minimally relaxing only the negotiable specifications while keeping all non-negotiable specifications strict.

For each negotiable formula $\varphi\in\Phi_S$, introduce a relaxation variable $\delta_\varphi\ge 0$ and relax its robustness constraint as
\begin{equation}
\rho_{\varphi}(x_{0:T}) \ge -\delta_\varphi, \quad \forall \varphi\in\Phi_S.
\label{eq:relaxed_robustness}
\end{equation}
Let $\delta\in\mathbb{R}_{\ge 0}^{|\Phi_S|}$ collect all relaxations. We then compute the least-violation modification of the negotiable specifications via
\begin{equation}
\begin{aligned}
\min_{u_{0:T-1},\,\delta \ge 0}\quad & \|\delta\|_1 \\
\text{s.t.}\quad
& x_{t+1} = f(x_t,u_t), \quad t=0,\dots,T-1,\\
& u_t \in \mathcal{U}, \quad t=0,\dots,T-1,\\
& \rho_{\varphi}(x_{0:T}) \ge 0, \quad \forall \varphi\in\Phi_H,\\
& \rho_{\varphi}(x_{0:T}) \ge -\delta_\varphi, \quad \forall \varphi\in\Phi_S.
\end{aligned}
\label{eq:min_relax}
\end{equation}
Denote an optimizer by $(u^\star,\delta^\star)$ and define the minimal relaxation level
\begin{equation}
\Delta_{\min} := \|\delta^\star\|_1.
\label{eq:delta_min}
\end{equation}
Equation~\eqref{eq:min_relax} characterizes the minimum aggregate relaxation required to recover feasibility while preserving strict satisfaction of $\Phi_H$. In particular, $\Delta_{\min}=0$ indicates that the original specification set is jointly satisfiable under the current context, whereas $\Delta_{\min}>0$ quantifies the smallest robustness violation (over negotiable formulas) needed to make the constrained MPC admit at least one feasible solution.

While \eqref{eq:min_relax} characterizes the minimum relaxation level needed for feasibility, the optimizer is generally non-unique: multiple pairs $(u,\delta)$ can attain the same $\Delta_{\min}$, yet allocate relaxation differently across $\Phi_S$ and induce different downstream behaviors. Next, we address this ambiguity by evaluating the consequences of these alternative feasibility-restoring resolutions.

\subsection{ Counterfactual, Value-Aware Tradeoff Refinement}
\label{subsec:stage2}

In this section, we compare alternative feasible conflict resolutions through their counterfactual consequences under the same context. Each feasible pair $(u,\delta)$ specifies (i) how negotiable formulas are relaxed and (ii) which control sequence is executed under those relaxed constraints. This defines a distinct hypothetical evolution of the system, whose consequences are scored by a vector of objectives.

Let
\begin{equation}
\mathbf{g}(u,\delta) = \big(g_1(u,\delta), \dots, g_m(u,\delta)\big)^\top
\label{eq:objective_vector}
\end{equation}
collect $m$ consequence objectives. We refine the feasibility-driven baseline from the previous section by seeking nondominated tradeoffs among these objectives within a relaxation budget:
\begin{equation}
\begin{aligned}
\min_{u_{0:T-1},\,\delta \ge 0}\quad & \mathbf{g}(u,\delta) \\
\text{s.t.}\quad
& x_{t+1} = f(x_t,u_t), \quad t=0,\dots,T-1,\\
& u_t \in \mathcal{U}, \quad t=0,\dots,T-1,\\
& \rho_{\varphi}(x_{0:T}) \ge 0, \quad \forall \varphi\in\Phi_H,\\
& \rho_{\varphi}(x_{0:T}) \ge -\delta_\varphi, \quad \forall \varphi\in\Phi_S,\\
& \|\delta\|_1 \in [\Delta_{\min},\, \Delta_{\min}+\alpha],
\end{aligned}
\label{eq:stage2_moo}
\end{equation}
where $\alpha\ge 0$ expands the admissible relaxation budget beyond the feasibility-restoring minimum. When $\alpha=0$, we compare alternatives that achieve the minimum total relaxation. When $\alpha>0$, it allows slightly larger relaxation to compare additional objective tradeoffs that may be preferred.

This formulation searches jointly over the relaxation allocation $\delta$ and the resulting control sequence $u$, and it evaluates tradeoffs in the space of consequences $\mathbf{g}(u,\delta)$ rather than in the space of relaxation variables. In particular, different allocations of the same relaxation budget can induce different trajectories and therefore different risk and performance outcomes, which are considered and evaluated explicitly.

\subsubsection{Autonomous-driving: dynamics and consequence objectives}
\label{subsubsec:av_instantiation}

We now instantiate the consequence objectives $\mathbf{g}(u,\delta)$ for autonomous driving, where conflicts naturally arise among safety requirements, traffic regulations, and task objectives. In this instantiation, the vehicle dynamics specify the MPC prediction model $f$, and $\mathbf{g}(u,\delta)$ is defined using risk-based metrics derived from predicted interactions with surrounding agents, optionally combined with task-performance terms (e.g., progress or comfort).

\textbf{Vehicle model.}
We model the ego vehicle using an acceleration-controlled kinematic bicycle model:
\begin{equation}
\begin{bmatrix}
\dot{p}_x \\
\dot{p}_y \\
\dot{\theta} \\
\dot{v}
\end{bmatrix}
=
\begin{bmatrix}
v \cos(\theta + \beta) \\
v \sin(\theta + \beta) \\
\dfrac{v}{l_r} \sin(\beta) \\
a
\end{bmatrix},
\end{equation}
where $(p_x,p_y)$ is planar position, $\theta$ is heading, $v$ is forward speed, $a$ is longitudinal acceleration, and $\delta$ is the front-wheel steering angle. The slip angle is
\begin{equation}
\beta = \tan^{-1}\!\left( \frac{l_r}{l_f + l_r} \tan(\delta) \right),
\end{equation}
with $l_f$ and $l_r$ denoting the distances from the vehicle center of mass to the front and rear axles, respectively.
Assuming small slip ($\cos\beta \approx 1$ and $\sin\beta \approx \beta$), we use the following control-affine approximation:
\begin{equation}
\underbrace{
\begin{bmatrix}
\dot{p}_x \\
\dot{p}_y \\
\dot{\theta} \\
\dot{v}
\end{bmatrix}
}_{\dot{x}}
=
\underbrace{
\begin{bmatrix}
v \cos \theta \\
v \sin \theta \\
0 \\
0
\end{bmatrix}
}_{f(x)}
+
\underbrace{
\begin{bmatrix}
0 & -v \sin \theta \\
0 & \;\, v \cos \theta \\
0 & \dfrac{v}{l_r} \\
1 & 0
\end{bmatrix}
}_{g(x)}
\underbrace{
\begin{bmatrix}
a \\
\beta
\end{bmatrix}
}_{u}.
\end{equation}
In implementation, we discretize and locally linearize this model to integrate with a MILP encoding of STL robustness constraints.

\textbf{Risk-based consequence objectives.}
To evaluate counterfactual consequences in this stage, we instantiate components of $\mathbf{g}(u,\delta)$ using a risk measure for each surrounding agent $j$:
\begin{equation}
R_j(u) \;=\; P_j(u) \times S_j(u) \times V_j,
\label{eq:risk_decomposition}
\end{equation}
where $P_j(u)$ is the probability of collision, $S_j(u)$ is the collision severity, and $V_j$ is the vulnerability of agent $j$.

The collision probability $P_j(u)$ is approximated via sampling. Let $x_{\mathrm{ego}}(t)$ denote the ego trajectory induced by the candidate control sequence $u$, and let $x_j^{(n)}(t)$ denote the $n$-th sampled trajectory of agent $j$, generated by perturbing the nominal prediction with Gaussian noise in velocity. The empirical collision probability is
\begin{equation}
P_j(u)
=
\frac{1}{N}
\sum_{n=1}^{N}
\mathbf{1}
\!\left[
\exists\, t :
d\!\left(x_{\mathrm{ego}}(t), x_j^{(n)}(t)\right)
\le d_{\mathrm{safe}}
\right],
\label{eq:prob_saa}
\end{equation}
where $d(\cdot,\cdot)$ is a distance metric (implemented using a box-based $L_\infty$ approximation) and $d_{\mathrm{safe}}$ is a safety threshold.

For sampled trajectories that collide, define the first contact time
\begin{equation}
t_j^{(n)}
=
\inf
\left\{
t
\;\middle|\;
d\!\left(x_{\mathrm{ego}}(t), x_j^{(n)}(t)\right)
\le d_{\mathrm{safe}}
\right\}.
\label{eq:first_contact}
\end{equation}
Let $v_{\mathrm{ego}}(t)$ and $v_j^{(n)}(t)$ denote velocity vectors at time $t$. Using the reduced mass
\begin{equation}
\mu_j
=
\frac{m_{\mathrm{ego}}\, m_j}
     {m_{\mathrm{ego}} + m_j},
\end{equation}
we compute the (unnormalized) severity as the sample mean of the reduced-mass--scaled relative speed at impact:
\begin{equation}
\tilde{S}_j(u)
=
\frac{1}{k}
\sum_{n=1}^{k}
\mu_j
\,
\left\|
v_{\mathrm{ego}}\!\left(t_j^{(n)}\right)
-
v_j^{(n)}\!\left(t_j^{(n)}\right)
\right\|_2,
\label{eq:severity}
\end{equation}
where $k$ is the number of colliding samples among $N$. We normalize $S_j(u)\in[0,1]$ via
\begin{equation}
S_j(u) = \frac{\tilde{S}_j(u)}{S_{\max}},
\end{equation}
which also facilitates setting meaningful grids for the $\varepsilon$-constraint bounds.

Finally, vulnerability captures susceptibility to injury independent of collision likelihood and impact severity. For an agent of type $\mathrm{type}_j$ with protection index $\kappa_{\mathrm{type}_j}\ge 0$, we define
\begin{equation}
V_j
=
\frac{1}{1 + \kappa_{\mathrm{type}_j}},
\end{equation}
ensuring $V_j\in(0,1]$. Pedestrians have higher $V_j$ due to minimal structural protection, whereas enclosed vehicles have lower values.
In practice, we construct $\mathbf{g}(u,\delta)$ by combining selected risk terms $\{R_j(u)\}$ and additional performance objectives as needed by the scenario.

\begin{algorithm}[t]
\caption{Pareto Approximation via the $\varepsilon$-Constraint Method}
\label{alg:pareto_epsilon}
\begin{algorithmic}[1]
\Require $\Delta_{\min}$, $\alpha$, objective vector $\mathbf{g}(u,\delta)$
\Ensure Approximate Pareto set $\mathcal{P}$ and Pareto front $\mathcal{F}$
\State Construct objective grids $\{E_\ell\}_{\ell=1}^{m}$ over attainable ranges (Sec.~\ref{subsubsec:pareto})
\State Initialize candidate set $\mathcal{C}\leftarrow\emptyset$
\For{$k=1$ to $m$}
    \For{each selection $\{\varepsilon_\ell\}_{\ell\neq k}$ with $\varepsilon_\ell\in E_\ell$}
        \State Solve \eqref{eq:epsilon_constraint} (optionally warm-start from the previous solve)
        \If{feasible}
            \State Add $(u,\delta,\mathbf{g}(u,\delta))$ to $\mathcal{C}$ \Comment{store consequences for dominance checks}
        \EndIf
    \EndFor
\EndFor
\State Compute $\mathcal{P}$ as the nondominated subset of $\mathcal{C}$ (w.r.t.\ $\mathbf{g}$)
\State Set $\mathcal{F} \leftarrow \{\mathbf{g}(u,\delta)\mid (u,\delta)\in\mathcal{P}\}$
\State \Return $\mathcal{P},\mathcal{F}$
\end{algorithmic}
\end{algorithm}

\subsubsection{Organizing counterfactual tradeoffs by nondominance}
\label{subsubsec:pareto}

To compare feasible conflict resolutions without committing to a single preference weighting, we adopt a dominance-based notion of optimality. We say that $(\bar u,\bar\delta)$ \emph{dominates} $(u,\delta)$ if $g_k(\bar u,\bar\delta)\le g_k(u,\delta)$ for all $k$ and the inequality is strict for at least one objective. A feasible solution is \emph{Pareto optimal} if it is not dominated by any other feasible solution. The Pareto set and its image in objective space provide a compact summary of achievable counterfactual tradeoffs.

Computing the Pareto set of \eqref{eq:stage2_moo} exactly is generally intractable, so we approximate it by generating a representative candidate set and pruning it by dominance. We use the standard $\varepsilon$-constraint construction, which produces candidates by solving a family of single-objective subproblems with explicit bounds on the remaining objectives.

\noindent\textbf{Remark 1 (Returned set is nondominated among explored candidates).}
Algorithm~\ref{alg:pareto_epsilon} generates a finite candidate set $\mathcal{C}$ by solving $\varepsilon$-bounded subproblems and then prunes $\mathcal{C}$ by dominance to obtain $\mathcal{P}$. By construction, every $(u,\delta)\in\mathcal{P}$ is nondominated within $\mathcal{C}$. When the $\varepsilon$ grids densely cover the attainable objective ranges and each subproblem is solved to global optimality, $\mathcal{P}$ provides a practical approximation to the Pareto set of \eqref{eq:stage2_moo}.

For each objective index $k\in\{1,\dots,m\}$, the $\varepsilon$-constraint subproblem is
\begin{equation}
\begin{aligned}
\min_{u_{0:T-1},\,\delta \ge 0}\quad & g_k(u,\delta) \\
\text{s.t.}\quad
& g_{\ell}(u,\delta) \le \varepsilon_{\ell}, \quad \forall \ell \neq k,\\
& x_{t+1} = f(x_t,u_t), \quad t=0,\dots,T-1,\\
& u_t \in \mathcal{U}, \quad t=0,\dots,T-1,\\
& \rho_{\varphi}(x_{0:T}) \ge 0, \quad \forall \varphi\in\Phi_H,\\
& \rho_{\varphi}(x_{0:T}) \ge -\delta_\varphi, \quad \forall \varphi\in\Phi_S,\\
& \|\delta\|_1 \in [\Delta_{\min},\, \Delta_{\min}+\alpha].
\end{aligned}
\label{eq:epsilon_constraint}
\end{equation}
For each $\ell\neq k$, the bound $\varepsilon_\ell$ is selected from a grid
$E_\ell = \{\varepsilon_\ell^1,\dots,\varepsilon_\ell^c\}$.
Sweeping $\{\varepsilon_\ell\}_{\ell\neq k}$ over these grids yields a set of feasible candidates $\mathcal{C}$, which we prune by dominance to obtain an approximate Pareto set $\mathcal{P}$ and the corresponding Pareto front
$\mathcal{F}=\{\mathbf{g}(u,\delta)\mid (u,\delta)\in\mathcal{P}\}$.

\begin{figure*}[t]
\centering
\setlength{\tabcolsep}{0pt}
\renewcommand{\arraystretch}{0}

\begin{tabular}{cccc}
\includegraphics[width=0.249\textwidth]{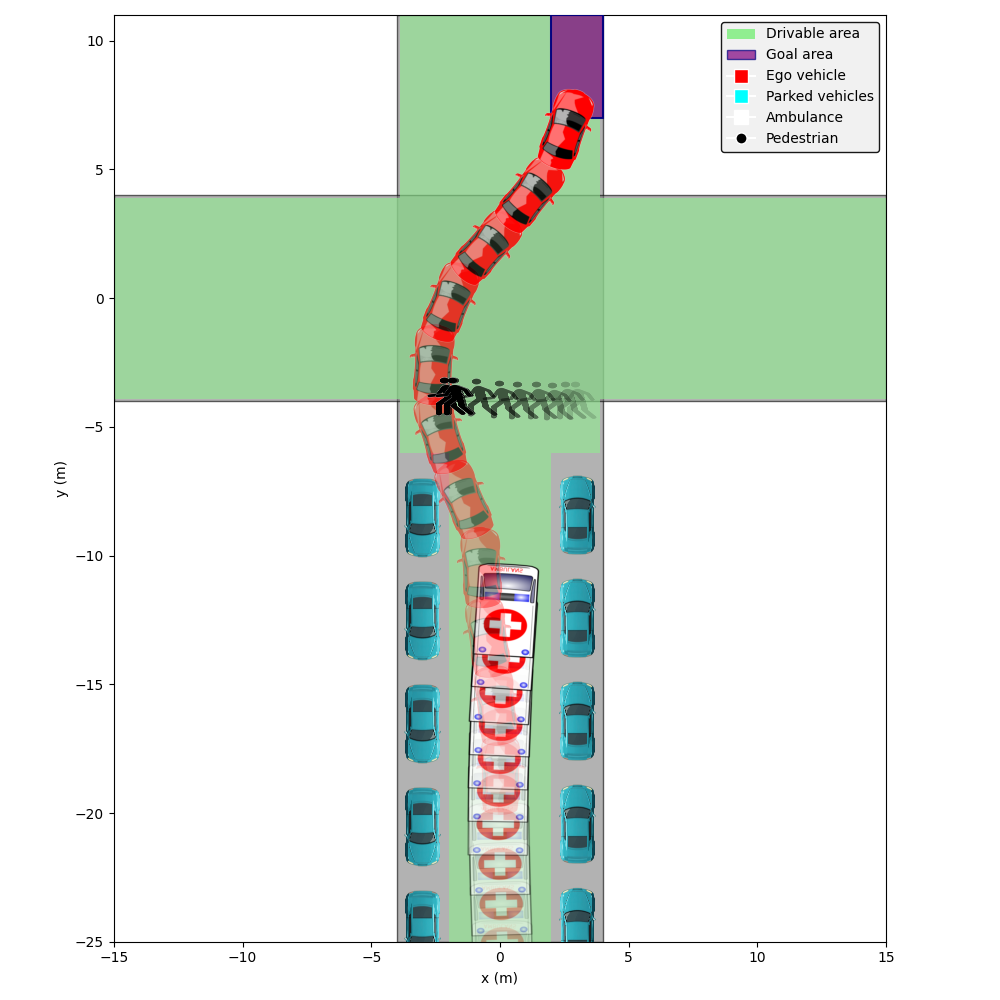} &
\includegraphics[width=0.249\textwidth]{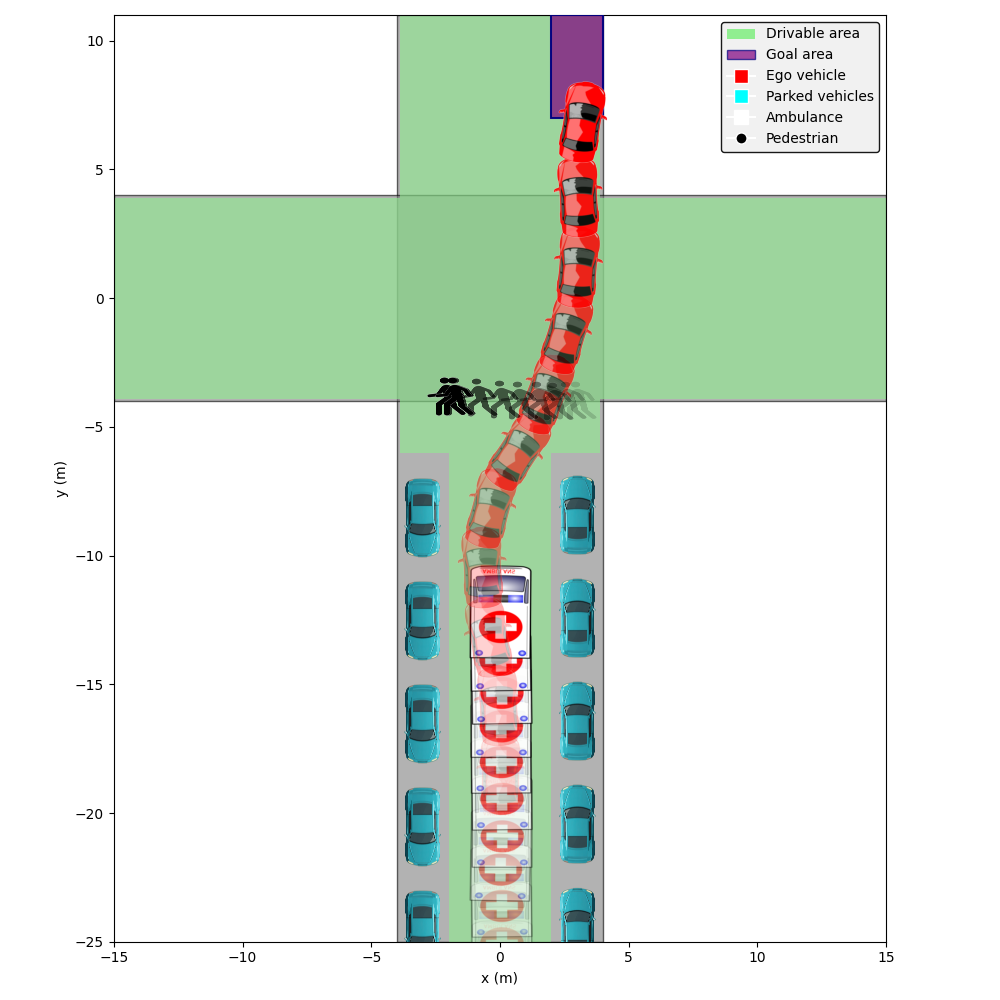} &
\includegraphics[width=0.249\textwidth]{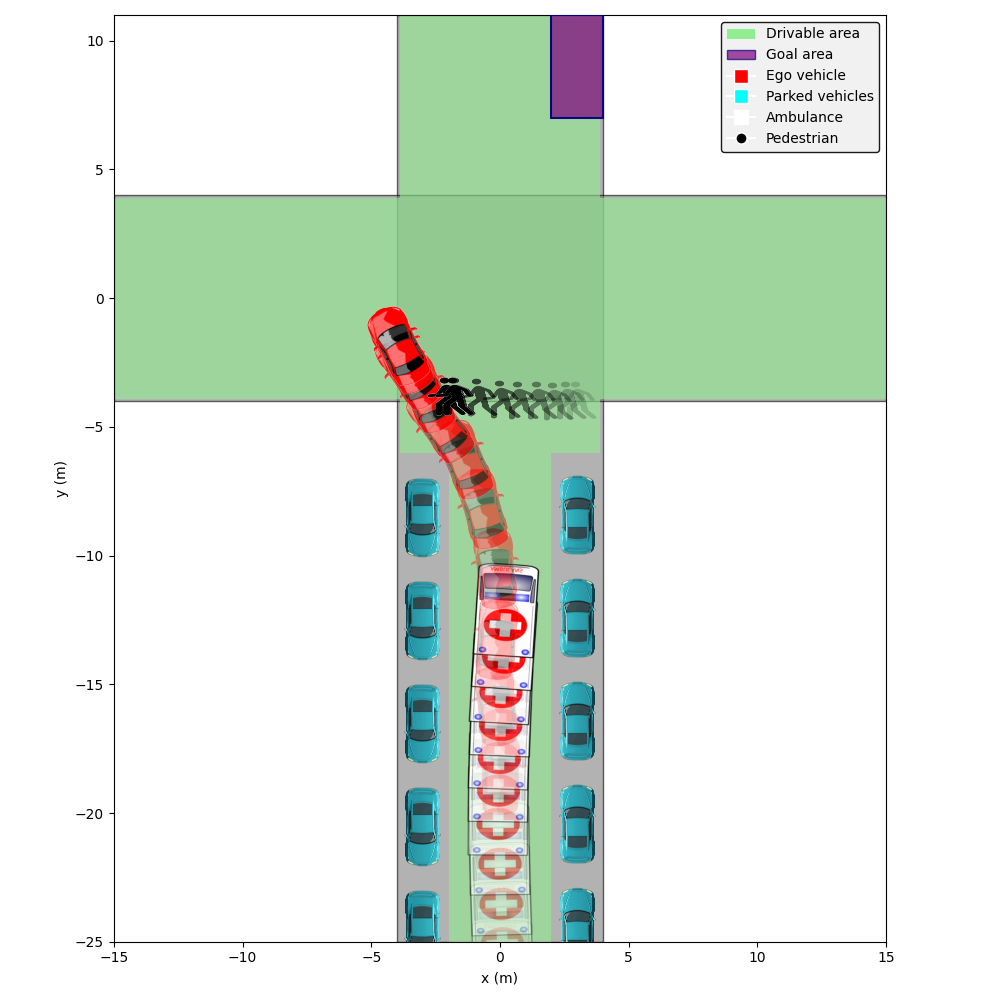} &
\includegraphics[width=0.249\textwidth]{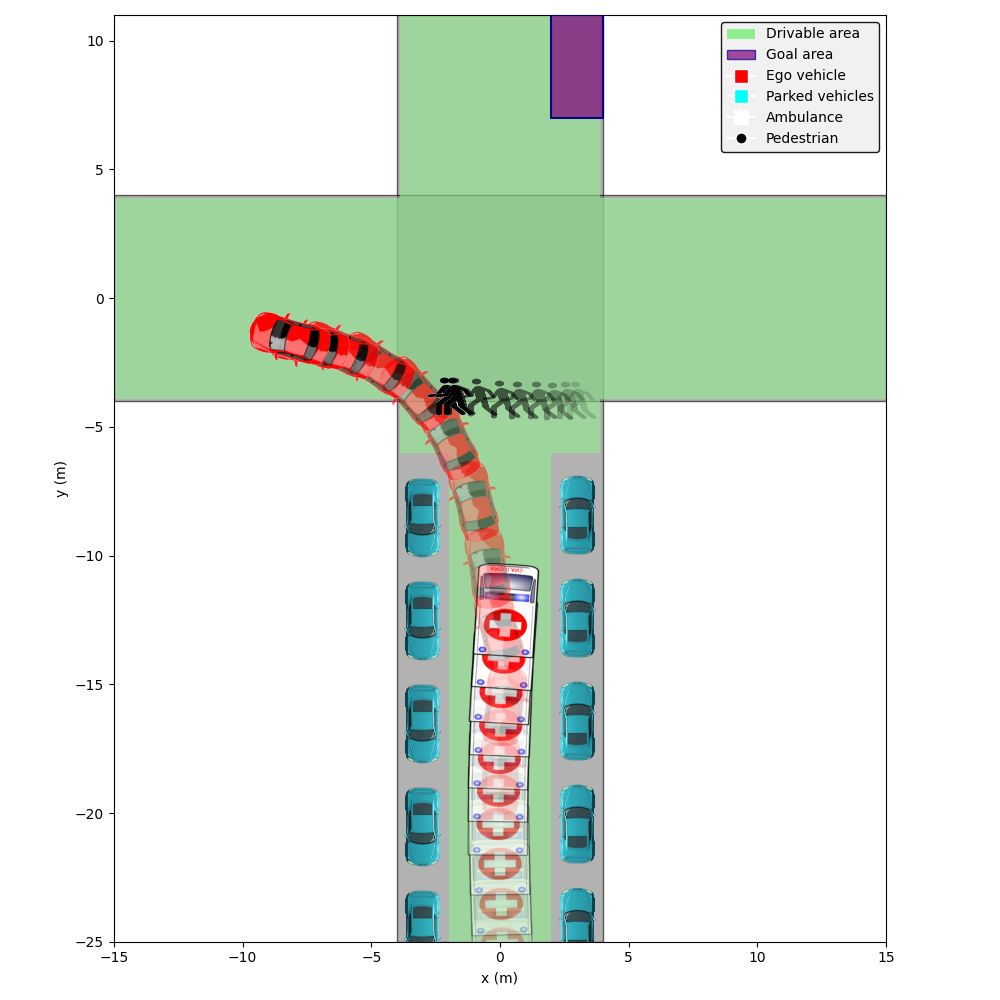} \\

\vspace{-2mm}(a) & \vspace{-2mm}(b) & \vspace{-2mm}(c) & \vspace{-2mm}(d)
\end{tabular}

\vspace{8mm}  

\caption{Ego-vehicle trajectory outcomes under competing STL rules. (a)-(c) illustrate feasible but \emph{dominated} solutions. (d) shows the selected \emph{nondominated} (Pareto-optimal) solution, which achieves a principled compromise among the risk objectives under the same observed scene.}

\label{fig:exp1_results}
\end{figure*}

Algorithm~\ref{alg:pareto_epsilon} summarizes the candidate generation and dominance filtering procedure. In the autonomous-driving setting, each $(u,\delta)\in\mathcal{P}$ corresponds to a distinct feasibility-restoring resolution of the competing rules in $\Phi_S$ that remains consistent with $\Phi_H$, together with the resulting ego trajectory. The Pareto front $\mathcal{F}$ therefore provides a structured set of counterfactual outcomes: moving along $\mathcal{F}$ reveals how reducing one risk imposed to a vehicle increases that imposed to a pedestrian. In the next section, we visualize these context-dependent safety tradeoffs and select a final control action according to a scenario-specific preference without collapsing objectives into fixed weights.


%% file: experiments.tex
\section{Experiments}

We evaluate the proposed two-stage conflict-resolution framework in two autonomous-driving scenarios where the negotiable STL specifications become conflicting in safety-critical moments. The first experiment shows how feasibility restoration alone can admit diverse qualitatively distinct behaviors that are not consequence-optimal, and the second one illustrates how the value-aware refinement stage changes relaxation allocation and yields a safer response under the same hard constraints.

In both experiments, we run MPC in a receding-horizon loop. At each control cycle, Stage~1 restores feasibility with respect to the conflicting specifications and obtain the minimum relaxation level $\Delta_{\min}$. Stage~2 then refines the control and relaxation by minimizing the risk function associated with each agent. From the Pareto front, we select the control with smallest deviation from a nominal controller.

\subsection{Exp1: Intersection conflict between emergency passage and pedestrian right-of-way}
\label{subsec:exp1}

\subsubsection{Problem setup}
The ego vehicle approaches an unsigned intersection and aims to reach a goal region across the intersection. An ambulance on an emergency mission approaches from behind. Because parked vehicles line both sides of the alley, the ego cannot safely pull over to yield. The ego therefore has an incentive to clear the intersection promptly to avoid obstructing the ambulance. As the ego is about to enter the intersection, a pedestrian suddenly starts crossing.

The non-negotiable requirement enforces that the ego must remain within the drivable region: 


\begin{equation}
\varphi_{\text{drivable}} = \mathbf{G}_{[0,T]}\big(x_{\text{ego}} \in \mathcal{R}_{\text{drivable}}\big).
\end{equation}
The negotiable specifications encode task completion and inter-agent safety requirements:
\begin{align}
\varphi_{\text{reach}} &= \mathbf{F}_{[0,T]}\big(x_{\text{ego}} \in \mathcal{R}_{\text{goal}}\big), \\
\varphi_{\text{ped safe}} &= \mathbf{G}_{[0,T]}\big(d(x_{\text{ego}},x_{\text{ped}})\ge d_{\text{safe}}\big), \\
\varphi_{\text{amb safe}} &= \mathbf{G}_{[0,T]}\big(d(x_{\text{ego}},x_{\text{amb}})\ge d_{\text{safe}}\big).
\end{align}
In this situation, advancing to reach the goal and avoid blocking the ambulance pushes the ego into the intersection, which may result in a collision with the pedestrian. Meanwhile, maintaining safe distance from the pedestrian may lead to a rear collision with the ambulance.

We use the following constants: $l_r=1.5$~m, $a_{\min}=-9$~m/s$^2$, $a_{\max}=4$~m/s$^2$, $\beta_{\min}=-0.2$~rad, $\beta_{\max}=0.2$~rad, $m_{\text{ego}}=1500$~kg, $m_{\text{ped}}=70$~kg, $m_{\text{amb}}=5000$~kg, $\kappa_{\text{ped}}=0.1$, $\kappa_{\text{veh}}=1.5$, $\kappa_{\text{amb}}=2.3$, $d_{\text{safe}}=2$~m, $\alpha=4$, horizon $T=2$~s, and time step $\Delta t=0.2$~s.

\subsubsection{Results and discussion}
Fig.~1 compares representative feasible trajectories that satisfy the feasibility-restoration requirement but differ in their risk objectives. Subplots (a)–(c) illustrate \emph{dominated} (non-Pareto) selections.
In Fig.~1(a), the ego performs a left maneuver to avoid the pedestrian while accelerating toward the goal. Although separation is maintained, the higher speed near the interaction region increases collision-severity risk. In Fig.~1(b), the ego instead steers right and continues forward without braking, producing a different spatial risk distribution and potentially increasing exposure in the intersection. In Fig.~1(c), the ego reduces speed to avoid the pedestrian but stops in a region aligned with the pedestrian's crossing direction. Additionally, abrupt deceleration increases the likelihood of an undesirable rear-end collision with the approaching ambulance.

In contrast, Fig.~1(d) shows a \emph{Pareto-optimal} selection that balances separation and velocity: the ego executes a gradual left turn while decelerating smoothly, reducing overall risk without introducing a strict increase in another evaluated risk component. This comparison illustrates why Stage~2 is necessary even after feasibility is restored: minimal relaxation prevents freezing, but Pareto-based refinement organizes counterfactual conflict resolutions and allows the controller to avoid strictly inferior choices.

Fig.~1(a)-(c) can be interpreted as counterfactual outcomes: under the same observed scene, if we were to select a different feasible candidate, the scene would evolve along those trajectories, leading to different risk distributions across the pedestrian, ambulance, and ego. Presenting these dominated alternatives alongside a Pareto-optimal selection provides an interpretable explanation of why a particular action was chosen, and what strictly inferior alternatives were rejected.


\subsection{Exp2: Out-of-control rear vehicle and emergency-lane borrowing}
\label{subsec:exp2}

\subsubsection{Problem setup}
The ego vehicle is stopped at a red light with two stationary vehicles ahead. In the opposite lane, a cyclist travels at moderate speed. Suddenly, an out-of-control vehicle approaches rapidly from behind, creating an imminent rear-end threat. The ego must react immediately to mitigate collision risk, potentially by borrowing an emergency lane, which conflicts with traffic rules.

The non-negotiable requirement constrains the ego to remain within the overall drivable region:
\begin{equation}
\varphi_{\text{drivable}} = \mathbf{G}_{[0,T]}\big(x_{\text{ego}} \in \mathcal{R}_{\text{drivable}}\big).
\end{equation}
The negotiable specifications include a traffic-rule constraint and two inter-agent safety constraints:
\begin{align}
\varphi_{\text{no emergency lane}} &= \mathbf{G}_{[0,T]}\big(x_{\text{ego}} \notin \mathcal{R}_{\text{emergency lane}}\big),\\
\varphi_{\text{cyclist safe}} &= \mathbf{G}_{[0,T]}\big(d(x_{\text{ego}},x_{\text{cyc}})\ge d_{\text{safe}}\big),\\
\varphi_{\text{rear vehicle safe}} &= \mathbf{G}_{[0,T]}\big(d(x_{\text{ego}},x_{\text{rear}})\ge d_{\text{safe}}\big).
\end{align}

We use the same parameter settings as in Exp1, and additionally set $m_{\text{rear}}=1500$~kg and $\kappa_{\text{rear}}=1.5$ for the rear vehicle.

\subsubsection{Results and discussion}
To demonstrate how Stage~2 improves the result, we compare:
(i) a trajectory produced by Stage~1 minimal relaxation alone, and
(ii) a Pareto-optimal, risk-aware refinement within the relaxation budget.
Fig.~2(e) shows that the minimal-relaxation controller resolves infeasibility by making a slight right turn that minimally enters the emergency-lane region, effectively paying the smallest relaxation necessary to restore feasibility. However, this choice prioritizes the magnitude of relaxation rather than the time-critical reduction of rear-end risk. 
Fig.~2(f) shows that the Pareto-refined controller instead accelerates and steers more toward the emergency lane to leave the collision region quickly, yielding a different relaxation allocation that improves the risk objectives. In Fig.~3, we compare the corresponding control inputs, which aligns with the observed vehicle behaviors.
Fig.~4 further illustrates that the Pareto-optimal solution allocates more relaxation to the emergency-lane rule beyond the minimum while keeping relaxations on the safety specifications at or near zero until necessary, reflecting that it priorities to minimize the risk objectives.

\begin{figure}[t]
\centering
\setlength{\tabcolsep}{0pt}
\renewcommand{\arraystretch}{0}
\begin{tabular}{cccc}
\includegraphics[width=0.249\textwidth]{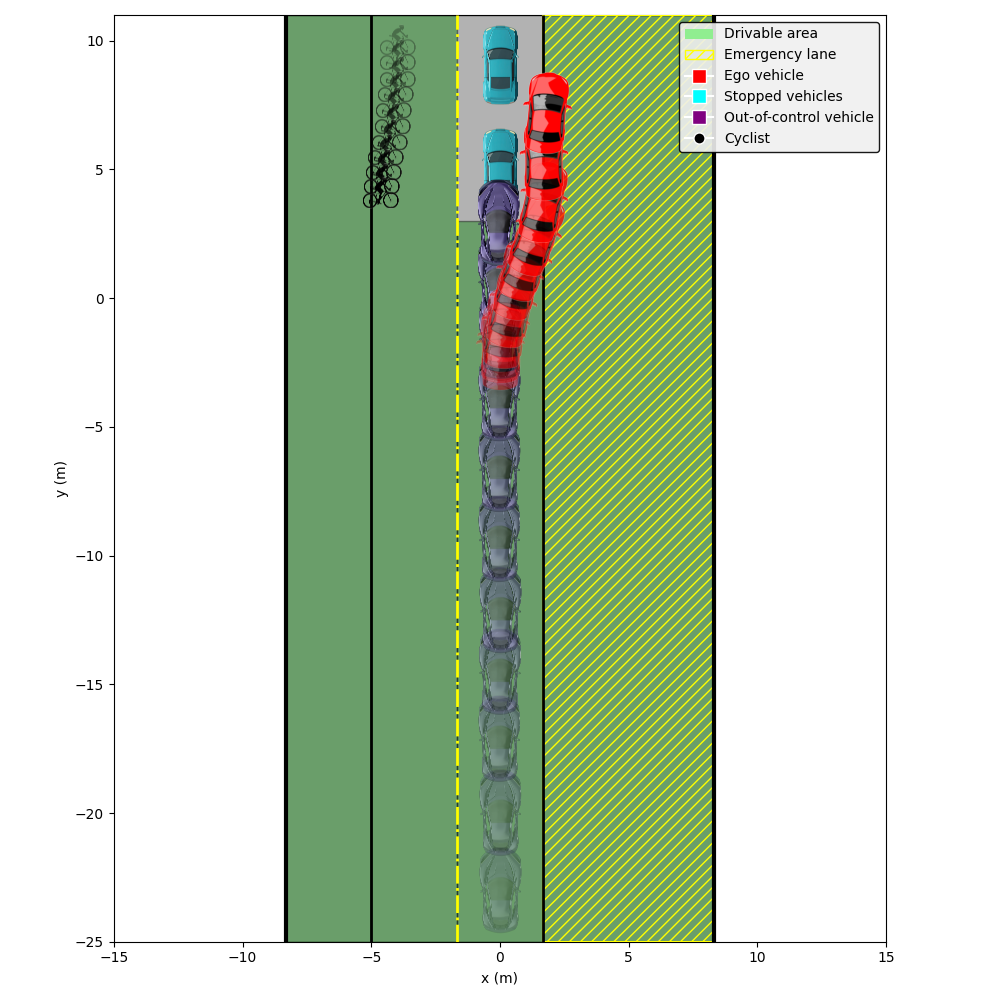} &
\includegraphics[width=0.249\textwidth]{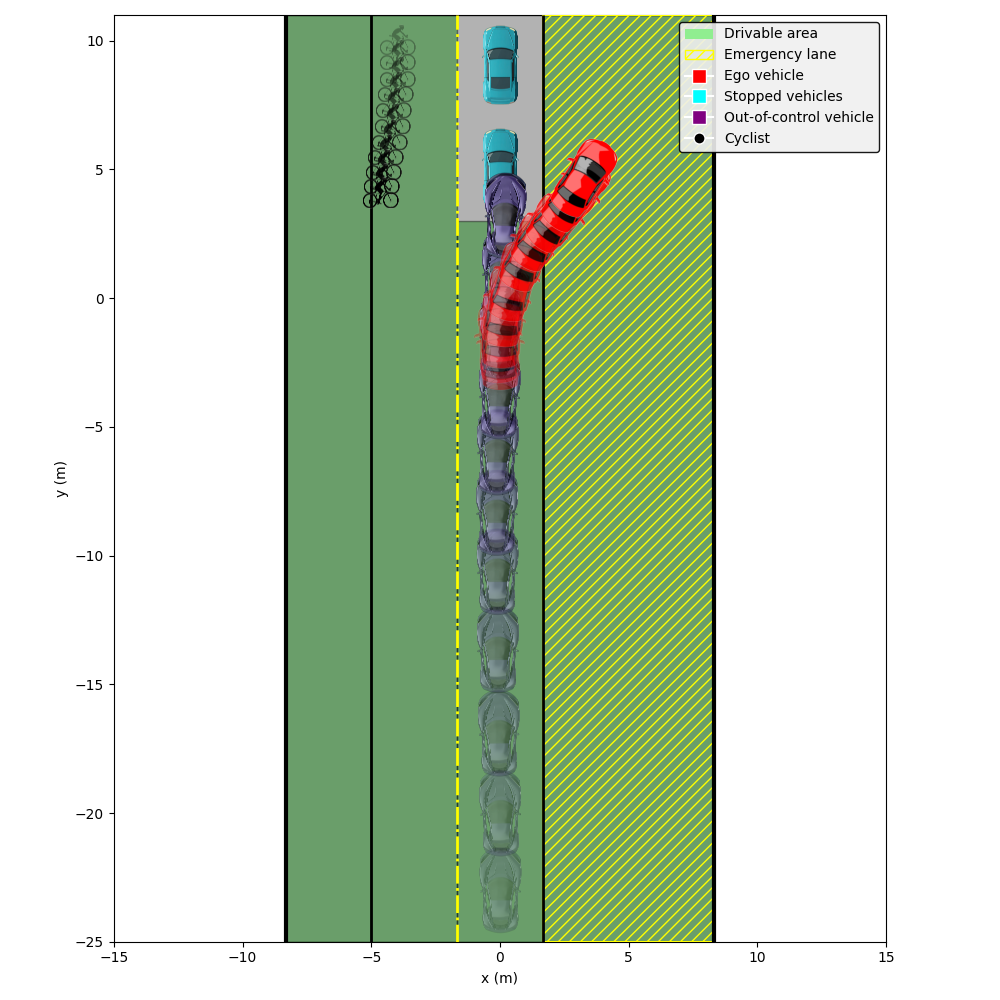} \\

\vspace{-2mm}(e) & \vspace{-2mm}(f)
\end{tabular}

\vspace{4mm}  

\caption{Stage-1 versus Stage-2 behavior in the same driving context. (e) shows the Stage-1 minimal-relaxation solution that restores feasibility with the smallest combined STL violation.  (f) shows a Stage-2 Pareto-optimal refinement that reallocates the admissible violation budget to improve risk objectives}
\label{fig:exp2_results}
\end{figure}

Exp2 also provides counterfactual interpretation, whereas under the same situation, one controller may minimize emergency-lane intrusion, but increase the risk of collision. Another may borrow the emergency lane more  to rapidly create separation from the collision spot, reducing rear-end risk while potentially increasing exposure relative to the oncoming cyclist. The Pareto front makes these alternatives explicit by showing which risk reductions necessarily require larger rule violations, and which candidates are strictly inferior and therefore rejected. This provides an interpretable rationale for the selected maneuver based on a nondominated compromise among safety and rule-compliance consequences in the given context.


\begin{figure}[t]
\centering
\setlength{\tabcolsep}{0pt}
\renewcommand{\arraystretch}{0}
\begin{tabular}{cccc}
\includegraphics[width=0.5\textwidth]{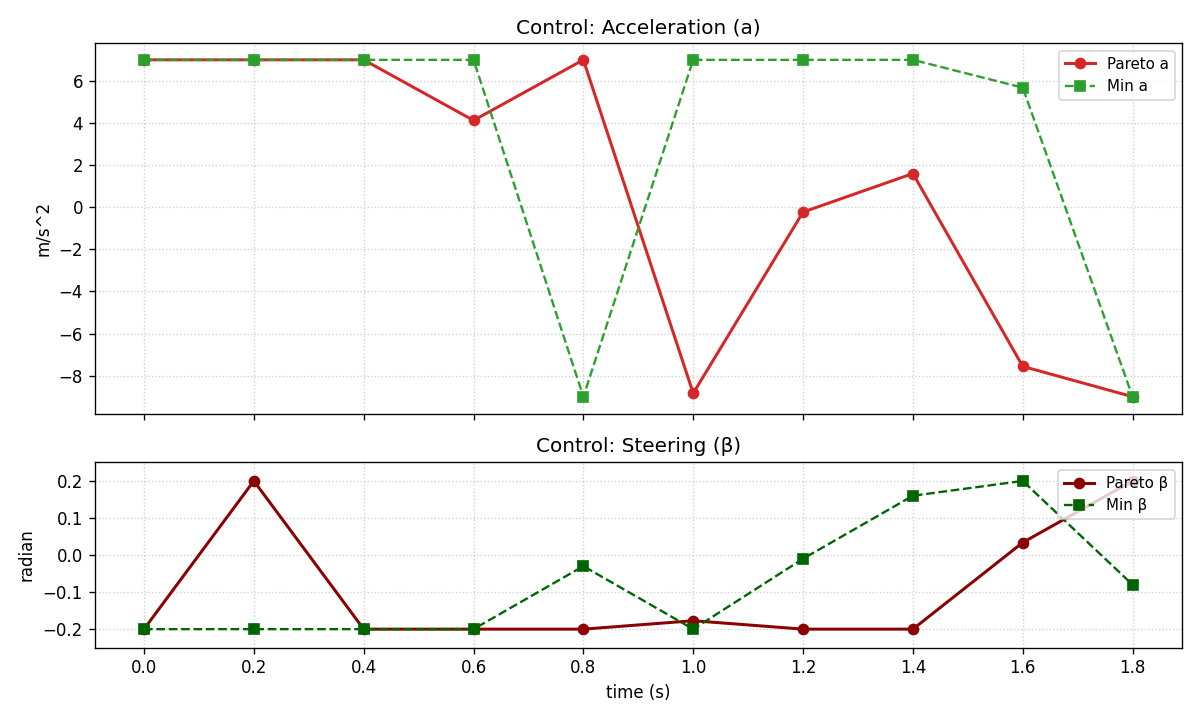}\\
\end{tabular}
\caption{Control sequences corresponding to Fig.~2.}
\label{fig:control}
\end{figure}


\begin{figure}[t]
\centering
\setlength{\tabcolsep}{0pt}
\renewcommand{\arraystretch}{0}
\begin{tabular}{cccc}
\includegraphics[width=0.5\textwidth]{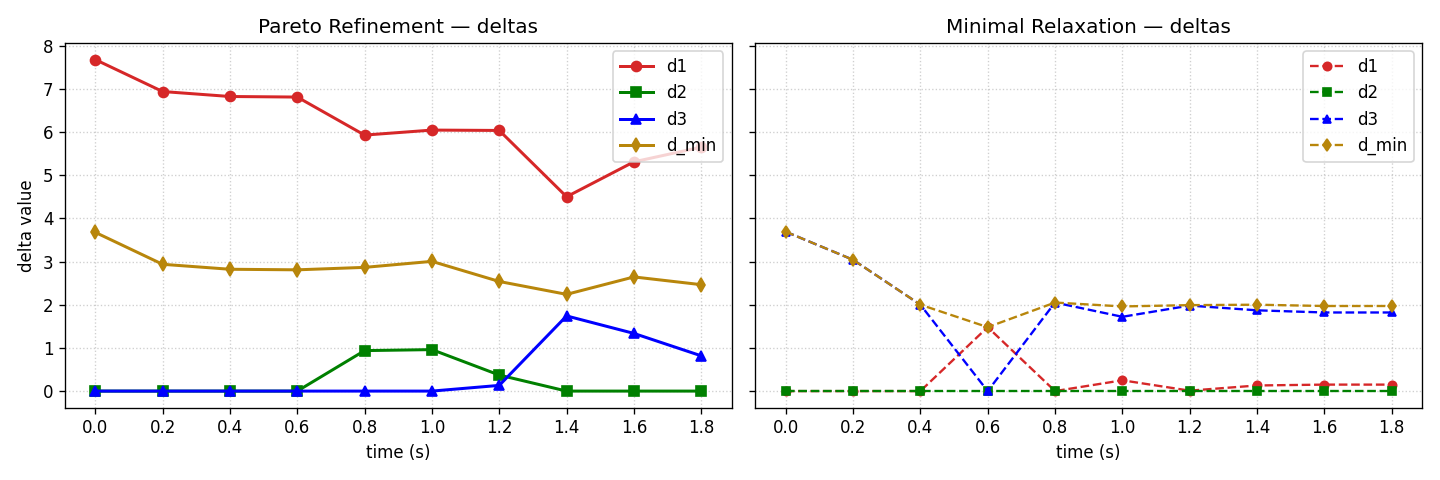}\\
\end{tabular}
\caption{Relaxation allocations across negotiable STL specifications, where d1, d2, d3 corresponds to the relaxation level of $\varphi_{\text{no emergency lane}}$, $\varphi_{\text{cyclist safe}}$ and $\varphi_{\text{rear vehicle safe}}$. $d_{min}$ indicates the minimal total relaxation to restore feasibility }
\label{fig:delta}
\end{figure}

%% file: conclusion.tex
\section{Conclusion}

We presented a two-stage, conflict-aware framework for STL-constrained MPC that maintains operation when runtime specifications become mutually inconsistent. Stage~1 restores feasibility by computing the minimum $L_1$ relaxation over negotiable STL formulas while keeping non-negotiable safety constraints strict, which prevents freezing behaviors of robots. Stage~2 treats each feasible relaxation allocation as a counterfactual decision and organizes the induced consequence tradeoffs via Pareto nondominance, using an $\varepsilon$-constraint construction to approximate the Pareto front over risk and performance objectives. Using autonomous driving as a motivating example, we show that feasibility restoration alone can admit many qualitatively different behaviors, while value-aware refinement selects nondominated resolutions that better manage risk under the same context. 

This work highlights the practical need to couple feasibility recovery with explicit tradeoff structure when deploying temporal-logic-based autonomy. Future directions include improving the efficiency of Pareto-front approximation for tighter real-time budgets, learning scenario-dependent selection policies over the nondominated set, and extending the framework to uncertainty models and multi-agent interaction settings.